\documentclass{article}
\usepackage[utf8]{inputenc}
\usepackage{pgfplots}
\usepackage{amsfonts}
\usepackage{algpseudocode}
\usepackage{bm}
\usepackage[square,sort,comma,numbers]{natbib}
\usepackage{algorithm}
\usepackage{subcaption}
\pgfplotsset{compat=1.18}
\usepackage[preprint]{neurips_2023}
\usepackage{booktabs}
\usepackage{amsmath}
\title{\textbf{TabADM: Unsupervised Tabular Anomaly Detection with Diffusion Models}}
\author{\textbf{Guy Zamberg$^1$}, \textbf{Moshe Salhov$^{2,4}$}, \textbf{Ofir Lindenbaum$^3$},  \textbf{Amir Averbuch$^2$} \\
$^1$School of Electrical Engineering, Tel Aviv University, Israel \\
$^2$School of Computer Science, Tel Aviv University, Israel \\
$^3$Faculty of Engineering, Bar-Ilan University, Israel \\
$^4$Playtika LTD, St. Herzllya, Israel
}
\begin{document}

\maketitle

\begin{abstract} 
Tables are an abundant form of data with use cases across all scientific fields. Real-world datasets often contain anomalous samples that can negatively affect downstream analysis. In this work, we only assume access to contaminated data and present a diffusion-based probabilistic model effective for unsupervised anomaly detection. Our model is trained to learn the density of normal samples by utilizing a unique rejection scheme to attenuate the influence of anomalies on the density estimation. At inference, we identify anomalies as samples in low-density regions. We use real data to demonstrate that our method improves detection capabilities over baselines. Furthermore, our method is relatively stable to the dimension of the data and does not require extensive hyperparameter tuning.
\end{abstract}

\section{Introduction} Anomaly detection, also known as outlier detection, involves identifying ``abnormal'' instances within datasets. These exceptional instances are called anomalies or outliers, while ``normal'' instances are known as inliers. In 1969, Grubbs \cite{GRUBBS} initially defined an outlier as "one that appears to deviate markedly from other members of the sample in which it occurs." Anomaly detection has numerous applications, such as fraud detection \cite{Fraud1,Fraud2}, network intrusion detection \cite{Intrusion1, Intrusion2}, medical diagnostics \cite{Medical1,irshaid2022histopathologic}, automatic explosion detection \cite{lindenbaum2016multi,bregman2021array} and social media \cite{Social1} to name some. 

To address the problem of anomaly detection, various methods have been proposed. The solutions can be classified into three settings: 1) Supervised, which requires a training set with labeled inliers/outliers but is limited due to the expensive data labeling. 2) Semi-supervised, which only requires pure single-class training data labeled as inliers without any outliers involved during training. 3) Unsupervised, which deals with completely unlabeled data mixed with outliers and does not require any data labeling for training. This paper deals with unsupervised anomaly detection, an approach that is widely applicable in practice due to the prevalence of unlabeled data. 

Existing unsupervised anomaly detection methods can be divided into different groups. The first group is subspace-based methods \cite{PCA,Subspace1, RDA, DAGMM,lindenbaum2021probabilistic}. The central assumption regarding these methods is that the normal data can be fully embedded in a lower-dimensional subspace. This assumption is not always valid and may constrain the range of applicable data distributions. Moreover, the performance of these methods depends heavily on the choice of hyperparameters used to define the subspace. 

Another family is based on data proximities or distances. Examples include K-Nearest Neighbors (KNN) \cite{KNN}, Local Outlier Factor (LOF) \cite{LOF}, and Cluster-Based Local Outlier Factor (CBLOF) \cite{CBLOF}. These methods define a data point as an outlier when its locality (or proximity)
is sparsely populated. Proximity-based methods are usually susceptible to the choice of distance measures. They also under-perform on high-dimensional data, where the curse of dimensionality causes distances to become less meaningful \cite{On_the_Surprising_Behavior_of_Distance, proximity_high_dim}. In addition, they typically require careful hyperparameter tuning, such as the number of neighbors or cluster size, which greatly influence their performance. 

Lastly, a group of probabilistic methods model the underlying distribution of the normal data and then identify data points exhibiting low probability under the model as potential anomalies. Particular methods \cite{GMM,Copod} limit the potential distributions by imposing assumptions on the interdependence of features or a specific parametric distribution. 
Additionally, some methods rely on Variational Autoencoders (VAEs) \cite{VAE} and Generative Adversarial Networks (GANs) \cite{GAN1,GAN2}. These methods may suffer from mode collapse, and hyperparameter tuning strongly influences their performance.

To overcome the above limitations, such as the reliance on prior assumptions that may restrict the generality of the data distribution, the challenging task of hyperparameter tuning, and the difficulty of coping with the curse of dimensionality in high-dimensional data, we introduce a novel approach from the probabilistic models family called Unsupervised Tabular Anomaly Detection with Diffusion Models (TabADM). On a high level, TabADM estimates the data distribution using a \textit{robust} diffusion generative model and then assigns an anomaly score to a new sample in correspondence to its probability of being generated by the model. Specifically, we rely on the training loss term to construct the anomaly score. To robustify the density estimation, we propose a sample rejection procedure to attenuate the influence of anomalies during training.

Our contributions are:
\begin{itemize}
\item Develop a method based on diffusion models for tabular anomaly detection. This method utilizes the stability property of diffusion models to avoid the challenge of hyperparameter tuning. Furthermore, it can be fully executed on a single laptop without requiring a GPU for most existing datasets.
\item Propose an anomaly rejection scheme to improve performance when the training set has outliers. We verify it on three different datasets and present scores improvement in all of them. 
\item Benchmark our method using multiple tabular datasets, demonstrating superior results with respect to two evaluation metrics compared with eleven popular detectors. In addition, our model significantly outperforms other competitors on high-dimensional datasets.   
%\item TabADM performs exceptionally well when on high-dimensional datasets. Diffusion models are specifically designed to handle high-dimensional datasets with complex probability distributions. As a result, TabADM has an inherent advantage when dealing with large and intricate datasets.
\end{itemize}

In this paper, we first provide a discussion of related work in the field of probabilistic models for anomaly detection in tabular data (Sec. \ref{RELATED_WORK}), followed by a description of our problem formulation and method (Sec. \ref{METHOD}). We then detail the experimental setup and report the results (Sec. \ref{EXPERIMENTS}). Finally, we discuss our findings and suggestions for future research directions (Sec. \ref{CONCLUSION}).

\section{Related Work} \label{RELATED_WORK}
Our method can be categorized under the family of probabilistic anomaly detection schemes. In this section, we first overview various probabilistic methods. Then, we discuss existing approaches for anomaly detection with diffusion models.

%In particular, our model is generative diffusion models based and generative relies on a likelihood function to evaluate the observations. Data points with a low probability of being generated by the model are assumed to be anomalies.

\paragraph{Parametric and non-parametric probabilistic methods.} Probabilistic models are usually categorized into two main groups, parametric and non-parametric. Methods that assume a specific parametric form of the underlying distribution are known as parametric methods. These methods aim to learn the parameters through a fitting process. A Common parametric framework is Gaussian Mixture Models based methods such as \cite{GMM}, in which the underlying distribution is modeled as a combination of multiple Gaussian distributions, and only the parameters of each Gaussian component are estimated.
In contrast, non-parametric methods do not assume any parametric model for the data. Some ``shallow'' non-parametric methods include Histogram-Based Outlier Score (HBOS) \cite{HBOS}, which uses a histogram to estimate the underlying distribution of the data, and Empirical Cumulative distribution based Outlier Detection (ECOD) \cite{ECOD}, which estimates the density using an empirical cumulative distribution of each feature independently. Following the revolution of deep neural networks, ``deep'' non-parametric methods have been developed. Such as Single-Objective Generative Adversarial Active Learning (SO-GAAL) \cite{SO_GAAL} that utilizes GANs as the primary generative model and active learning to enhance detection performance. More recently, \cite{rozner2023anomaly} proposed variance stabilized density estimation for anomaly detection implemented using an autoregressive model.

\paragraph{Diffusion models for anomaly detection} 
Diffusion models \cite{DDPM} are a class of generative models that are used in many applications such as image generation \cite{Diffusion_Beat_GANs}, video generation \cite{Video_Diffusion_Models}, text-to-image generation \cite{Text_To_Image_Diffusion, Text_To_Image_Diffusion1}, semantic segmentation \cite{Semantic_Diffusion,Semantic_Diffusion1} and waveform signal processing \cite{Waveform_Diffusion}. Diffusion models have also been utilized in anomaly detection tasks. For instance, some methods \cite{Diffusion1,Diffusion2} focus on identifying anomalous regions within images, while others like \cite{video_diffusion} detect anomalous frames in videos. However, to the best of our knowledge, no existing methods for detecting anomalies in tabular data employ diffusion models.

\section{Method}\label{METHOD}
We begin by presenting the problem formulation for unsupervised anomaly detection. Then, we explain our proposed approach with a brief theoretical review of diffusion models. Lastly, we describe the algorithm and the network architecture.

\subsection{Problem Formulation}
\paragraph{Setup.} We follow the setup given in \cite{Latent_Outlier_Exposure} for the problem of unsupervised anomaly detection on tabular data. 
Suppose we have a tabular dataset $\bm{S}\in\mathbb{R}^{n \times d}$ consisting of $n$ samples $\mathbf{x}_i$ ($i=1,2,...,n$) with $d$ dimensions. Each sample $\mathbf{x}_i$ could either be a ``normal'' sample drawn from the data density distribution $q(\mathbf{x})$ or an ``anomaly'' drawn from an unknown corruption process. We also assume that anomaly samples are located in low-density regions. Our goal is to train an anomaly classifier $M$ on $\bm{S}$ and then, given a new data sample that is not part of $\bm{S}$, we want to assign an anomaly score indicating the degree to which it is anomalous (higher score means it more likely to be an anomaly).

\paragraph{Proposed Approach.} Following probabilistic anomaly detection approach, we train $M$ on contaminated $\bm{S}$ to model the density $q_S(\mathbf{x})$. Assuming that anomaly samples are located in low-density regions, we approximate that $q_S(\mathbf{x})=q(\mathbf{x})$. However, we take into account that the presence of anomalies has a detrimental effect on the modeling process. Therefore, we rely on the training loss to assign an anomaly score for an unseen data sample at inference. As demonstrated in the next paragraph, the loss is based on the log-likelihood of the model given the training data. Samples with low probability density under the learned distribution $q(\mathbf{x})$ are more likely to be anomalies and result in high loss values. Hence it can serve as a quantitative measure of abnormality. 
Following the success of diffusion models in generative modeling, we present a diffusion architecture to model $q(\mathbf{x})$. We now provide a concise overview of the diffusion framework.
 
\paragraph{Density modeling with diffusion models.}\label{Density_Modeling_with_DM} We briefly introduce the theory of diffusion models
mentioned in \cite{DDPM}. We begin by defining the data distribution $\mathbf{x}_0 \sim q(\mathbf{x}_0)$, where $\mathbf{x}_0 \in \mathbb{R}^d$. We fix a Markov chain to a noising process in which Gaussian noise is gradually added to $\mathbf{x}_0$ through $T$ consecutive diffusion steps, producing latent variables $\mathbf{x}_1,...,\mathbf{x}_T$ of noisy samples with the same dimensionality as $\mathbf{x}_0$. Particularly, for a noising variance schedule $\beta_1,...,\beta_T$:
\begin{equation*}
q\left(\mathbf{x}_{1: T}|\mathbf{x}_0\right):=\prod_{t=1}^T q\left(\mathbf{x}_t|\mathbf{x}_{t-1}\right), \quad q\left(\mathbf{x}_t|\mathbf{x}_{t-1}\right):=\mathcal{N}\left(\mathbf{x}_t ; \sqrt{1-\beta_t} \mathbf{x}_{t-1}, \beta_t \mathbf{I}\right).
\end{equation*}
A notable property regarding $q(\mathbf{x}_t|\mathbf{x}_o)$ is that it can be expressed as Gaussian distribution. Let $\alpha_t := 1-\beta_t$ and $\alpha_t:=\prod_{s=1}^t \alpha_s$: 
\begin{equation}
\label{q(x_t|x_0)}
q(\mathbf{x}_t|\mathbf{x}_0)=\mathcal{N}(\mathbf{x}_t ; \sqrt{\bar{\alpha}_t} \mathbf{x}_0,(1-\bar{\alpha}_t) \mathbf{I}),
\end{equation}
Hence:
\begin{equation}
\label{x_t}
\mathbf{x}_t=\sqrt{\bar{\alpha}_t} \mathbf{x}_0+\sqrt{1-\bar{\alpha}_t}\mathbf{\epsilon}, \mathbf{\epsilon} \sim \mathcal{N}(0, \mathbf{I}).
\end{equation}
Using Bayes theorem on Eq. \ref{q(x_t|x_0)}:
\begin{equation}
\label{q_x_t-1}
q\left(\mathbf{x}_{t-1}|\mathbf{x}_t, \mathbf{x}_0\right)=\mathcal{N}(\mathbf{x}_{t-1} ; \tilde{\mu}_t\left(\mathbf{x}_t, \mathbf{x}_0\right), \tilde{\beta}_t \mathbf{I}),
\end{equation}
\begin{equation*}
\textrm{where} \quad \tilde{\boldsymbol{\mu}}_t\left(\mathbf{x}_t, \mathbf{x}_0\right):=\frac{\sqrt{\bar{\alpha}_{t-1}} \beta_t}{1-\bar{\alpha}_t} \mathbf{x}_0+\frac{\sqrt{\alpha_t}\left(1-\bar{\alpha}_{t-1}\right)}{1-\bar{\alpha}_t} \mathbf{x}_t, \quad \tilde{\beta}_t:=\frac{1-\bar{\alpha}_{t-1}}{1-\bar{\alpha}_t} \beta_t.
\end{equation*}
We aim to learn the data distribution $q(\mathbf{x}_0)$. We define distribution $p_\theta(\mathbf{x}_0)$ towards this goal. Since $q(\mathbf{x}_t|\mathbf{x}_{t-1})$ is Gaussian and if $\beta_t$ is small for all $t$, then $q(\mathbf{x}_{t-1}|\mathbf{x}_{t})$ is also Gaussian. Thus we can approximate $q(\mathbf{x}_{t-1}|\mathbf{x}_{t})$ using a neural network:
\begin{equation}
\label{p_theta_t-1}
p_{\theta}(\mathbf{x}_{t-1}|\mathbf{x}_t)=\mathcal{N}(\mathbf{x}_{t-1} ; \tilde{\mu}_\theta(\mathbf{x}_t, t), \Sigma_\theta(\mathbf{x}_t,t).
\end{equation}
Training the model such that $p_\theta(\mathbf{x}_0)$ estimates $q(\mathbf{x}_0)$, we optimize variational lower bound on the log likelihood:
\begin{equation}
\label{L_vlb}
L_{vlb} := L_0 + L_1 +...+ L_T,
\end{equation}
\begin{equation*}
\label{L_0}
L_0 := -\log{p}_\theta(\mathbf{x}_0|\mathbf{x}_1),
\end{equation*}
\begin{equation*}
\label{L_t-1}
L_{t-1} := D_{KL}(q(\mathbf{x}_{t-1}|\mathbf{x}_t,\mathbf{x}_0)||p_\theta(\mathbf{x}_{t-1}|\mathbf{x}_t),
\end{equation*}
\begin{equation*}
\label{L_T}
L_T := D_{KL}(q(\mathbf{x}_t|\mathbf{x}_0)||p(\mathbf{x}_T)).
\end{equation*}
%In order to minimize \ref{L_vlb}, we need to model 
Ho et al. \cite{DDPM} found out that objective loss (\ref{L_vlb}) can be simplified based on equivalency of  (\ref{q_x_t-1}) and (\ref{p_theta_t-1}) to the sum of mean squared errors between $\mathbf{\epsilon}$ and $\bm{\epsilon}_{\theta}(\mathbf{x}_t,t)$:
\begin{equation}
\label{L_simple}
L_{simple}(\theta) := \mathbb{E}_{t,\mathbf{x}_0,\mathbf{\epsilon}}[||\mathbf{\epsilon}-\mathbf{\epsilon}_\theta(\mathbf{x}_t,t)||_2^2].
 \end{equation}
More specifically, the model $\mathbf{\epsilon}_\theta(\mathbf{x}_t,t)$ is trained to predict the true noise $\mathbf{\epsilon}$ by minimizing the simplified objective loss (\ref{L_simple}). Each sample $\mathbf{x}_t$ is produced using Eq. (\ref{x_t}), by randomly drawing $\mathbf{x}_0$, $t$ and $\mathbf{\epsilon}$. 

\subsection{TabADM} The TabADM algorithm is composed of two sequential components, namely \textit{train} and \textit{inference}. In the \textit{training} phase, the model estimates the data distribution $q(\mathbf{x})$ of the training data. In addition, we include an anomaly rejection scheme during training to minimize the influence of existing anomalies in the data. At \textit{inference}, an anomaly score is assigned for each sample in the test data based on a summation of loss values at each diffusion timestep. These parts will be described in detail in Sec. \ref{sec:Train} and \ref{sec:Inference}. We conclude this part by presenting our architecture in Sec. \ref{ARCHITECTURE}.

\subsubsection{Train}\label{sec:Train} Algorithm \ref{Training} describes the \textit{train} part of TabADM algorithm. We train a model $\bm{\epsilon}_{\theta}(\mathbf{x}_t,t)$ to estimate the density $q(\mathbf{x})$ of the training data $\textbf{S} \in \mathbb{R}^{n \times d}$. As outlined in section \ref{Density_Modeling_with_DM}, the estimation of $q(\mathbf{x})$ involves the minimization of the objective loss (Eq. \ref{L_simple}) through a well-defined procedure. Specifically, the data is first normalized to be in the $[-1,1]$ interval, and a loop over $e$ steps is executed. At each step, a $k$-sample batch $\mathbf{x}_0$ is drawn from $\bm{S}$. In addition, a Gaussian noise 
$\bm{\epsilon}$ and a timesteps array $\bm{t}$ with $k$ copies of a randomly picked timestep $t$ are created to generate $\mathbf{x}_t$ according to Eq. \ref{x_t}. The model $\bm{\epsilon}_{\theta}(\mathbf{x}_t,\bm{t})$ estimates the true noise $\bm{\epsilon}$ and the loss (Eq. \ref{L_simple}) is calculated.  

\paragraph{Anomaly rejection scheme} To reduce the impact of potential anomalies $\bm{S}$, we utilize the loss function to estimate the probability that a sample is abnormal. We introduce the function $last_{k-m}(loss)$, which sorts the loss values in a batch of $k$ samples in descending order and keeps only the last $k\!-\!m$ values. Stochastic gradient descent (SGD) is applied using the $last_{k-m}(loss)$ to conduct the train iteration.

% Train algorithm
\begin{algorithm}[H]
    \caption{Train}\label{Training}
    \textbf{Input:} train data $\textbf{S} \in \mathbb{R}^{n \times d}$, batch size $k \in \mathbb{N}$, train steps $e \in \mathbb{N}$, rejection samples $m \in \mathbb{N}$, diffusion timesteps $T \in \mathbb{N}$
    \begin{algorithmic}[1]
        \State Normalize \textbf{S}
        \For{$i=1$ to $e$}
            \State Sample $\mathbf{x}_0\in \mathbf{S}$ \hfill\Comment{$\mathbf{x}_0 \in \mathbb{R}^{k \times d}$}
            \State Sample $\bm{\epsilon} \sim \mathcal{N}_{k \times d}(0, I)$ 
            \State Sample $t \in \mathcal{U}(\{1,...,T\})$
            \State Create array $\bm{t}$ with $k$ copies of $t$
            \State $\mathbf{x}_t = \sqrt{\bar{\alpha}_t}\mathbf{x}_0+\sqrt{1-\bar{\alpha}_t}\bm{\epsilon}$ \hfill\Comment{Eq. \ref{x_t}}
            \State $loss = ||\bm{\epsilon}-\bm{\epsilon}_{\theta}(\mathbf{x}_t,\bm{t})||^2_2$ \hfill\Comment{$loss \in \mathbb{R}^{k}$}
            \State SGD($last_{k-m}(loss)$)
        \EndFor
    \end{algorithmic}
\end{algorithm}

\subsubsection{Inference}\label{sec:Inference} 
Algorithm \ref{Inference} describes the \textit{inference} part of TabADM, which generates anomaly scores $\bm{O} \in \mathbb{R}^{k}$ to each sample in test data $\textbf{S} \in \mathbb{R}^{k \times d}$. To begin, we normalize $\bm{S}$ to the $[-1,1]$ interval according to the train data. In addition, we initialize the output anomaly scores array $\bm{O}$ with $k$ zeros and generate a Gaussian noise matrix $\bm{E} \sim \mathcal{N}_{T \times d}(0, I)$. For each sample in $\bm{S}$, a sequence $({\textbf{x}_t})_{t=1}^{T}$ of noisy data samples is generated, where each $\mathbf{x}_t$ is created from timestep $t$ and noise $\bm{E}_t$ (Eq. \ref{x_t}). The total loss for each sample is computed by summing the loss values across all timesteps, and it is stored in the corresponding sample entry in $\bm{O}$.

% Inference algorithm
\begin{algorithm}[H]
    \caption{Inference}\label{Inference}
    \textbf{Input:} test data $\bm{S} \in \mathbb{R}^{k \times d}$, diffusion timesteps $T \in \mathbb{N}$\\
    \textbf{Output:} Anomaly scores $\bm{O} \in \mathbb{R}^{k}$
    \begin{algorithmic}[1]
        \State Normalize $\bm{S}$ according to train data
        \State Initiate zeros array $\bm{O}$ of size $k$
        \State Initiate $\bm{E} \sim \mathcal{N}_{T \times d}(0, I)$
        \hfill\Comment{$\bm{E} \in \mathbb{R}^{T \times d}$}
        \For{$i=1$ to $k$}
            \State Pick $\mathbf{x}_0= \bm{S}_{i}$ 
            \For{$t=1$ to $T$}
                \State $\mathbf{x}_{t} = \sqrt{\bar{\alpha}_t}\mathbf{x}_0+\sqrt{1-\bar{\alpha}_t}\bm{E}_t$
                \hfill\Comment{Eq. \ref{x_t}}
                \State $ loss = ||\bm{E}_t -\bm{\epsilon}_{\theta}(\mathbf{x}_t,t)||^2_2$
                \State $O_i \mathrel{+}= loss$
            \EndFor
        \EndFor
        \State \textbf{Return} $\bm{O}=\{O_1,...,O_k\}$
    \end{algorithmic}
\end{algorithm}

\subsubsection{Architecture} \label{ARCHITECTURE}
\begin{figure}[H]
  \vskip -0.2 in
\centering
\includegraphics[width=1\textwidth]{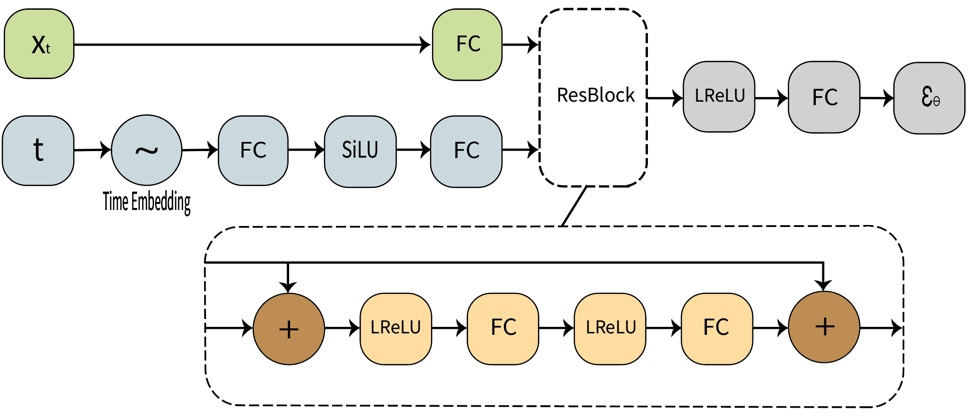}
\caption{Proposed architecture for anomaly detection on tabular data. The model receives noisy sample $\bm{x}_t$ and time step $t$ that are fed forward to the ResBlock. The output of the ResBlock propagates through the Leaky-ReLU activation function followed by the FC layer to create the noise estimation of the real noise component in $\mathbf{x}_t$.}
\label{fig:arcitecture}
\end{figure}

Our model $\bm{\epsilon}_\theta(\mathbf{x}_t,t)$ is a variation of ResNet architecture for tabular data \cite{Revisiting_Deep_Learning_Models_for_Tabular_Data} with the utilization of relevant components from U-Net model used in DDPM \cite{DDPM}. Specifically, we use a time embedding block defined by the Transformer sinusoidal position embedding \cite{Attention_Is_All_You_Need} and a single residual block (ResBlock) to combine the feature vectors of the time-step $t$ and the noisy sample $\mathbf{x}_t$. The sizes of the time embedding block and the fully connected (FC) layers are defined as hyperparameters (See Tab. ~\ref{table:hyperparameters list}). We use SiLU and Leaky-ReLU with a negative slope of 0.2 as activation functions. Fig. \ref{fig:arcitecture} describes the block diagram of our architecture. 

\section{Experiments} \label{EXPERIMENTS}
\paragraph{Datasets.} We use 32 anomaly detection datasets from the ADBench repository \cite{ADBench} in this study (Appx. Tab. \ref{table:datasets list}). Of these, 28 are real-world datasets, and the rest are extracted data-embedding representations of pre-trained models from the fields of computer vision (CV) and natural language processing (NLP). Specifically, the CV datasets include \textit{FashionMNIST} and \textit{SVHN}, for which both \textit{BERT} and \textit{RoBERTa} versions are utilized, and we randomly select the first class (out of 10 existing) for testing. The NLP datasets include \textit{Amazon} and \textit{Yelp}, and both \textit{ViT} and \textit{ResNet} versions are employed. In addition, due to convergence failure in some of the baselines, we stratified truncate \textit{Census} to $50K$ samples, i.e., we maintain the original anomaly ratio post truncation.

\paragraph{Baseline methods and hyperparameters Settings.} We evaluate TabADM against eleven outlier detectors. Among them, nine are leading detectors from ADBench \cite{ADBench} with a wide variety and two recent NN based methods. The competitors from ADBench are $k$ Nearest Neighbors (KNN) \cite{KNN}, Local Outlier Factor (LOF) \cite{LOF}, One-Class Support Vector Machines (OCSVM) \cite{OCSVM}, PCA-based Outlier Detector (PCA) \cite{PCA}, Clustring-based Local Outlier Factor (CBLOF) \cite{CBLOF}, Isolation Forest (IForest) \cite{IF}, Copula Based Outlier Detector (COPOD) \cite{Copod}, Histogram-based Outlier Detector (HBOS) \cite{HBOS} and Empirical Cumulative Distribution-based Outlier Detector (ECOD) \cite{ECOD}.
We use PyOD \cite{PyOD} anomaly detection python package for implementation of baseline methods and use their default PyOD\footnote{https://pyod.readthedocs.io/en/latest/pyod.html} configuration for a fair comparison\footnote{For CBLOF, we use $n_{clusters}=9$ due to convergence failure in some of the datasets using the default settings.}.
Additionally, we include GOAD by Bergman et al. \cite{GOAD} and NeuTraL AD (referred to as NeuTraL) by Qiu et al. \cite{NeuTral_AD}. These methods have recently demonstrated impressive results on tabular data. We adopt the \textit{kdd} based configuration for both methods for all experiments. The default hyperparameters we use for the training of TabADM are summarized in Appx. Tab. \ref{table:hyperparameters list}.

\paragraph{Results} We use the Area Under Receiver Operating Characteristic Curve (AUCROC) and Average Precision (AP) as evaluation metrics. We use a MacBook Pro laptop with M1, 16 GB of memory, and without GPU for all experimental runs.

In the \textbf{first part}, we follow the ADBench \cite{ADBench} experiment settings and use random stratified sampling to divide the data into 70\% for training and 30\% for testing. We repeat this process five times and report the average scores\footnote{For CV and NLP datasets, we report the average score of the two different versions.}. In addition, to evaluate the performance of our method in high-dimensional data, we sort the 32 datasets in ascending order according to their dimensions and define the parameter $\tau$ as the percentile value corresponding to the dimensions. For each value of $\tau$, we partition the datasets into subgroups based on $\tau$, where each subgroup consists of datasets with dimensions greater than $\tau$. For example, when $\tau=10$, we form a group comprising the top 90\% datasets with the highest number of variables. We calculate the average AUCROC and AP ranks for each sub-group for each method. We plot the values of the average ranks of both AUCROC and AP as a function of $\tau$.

\begin{figure}[H]
\centering
\includegraphics[width=1\textwidth]{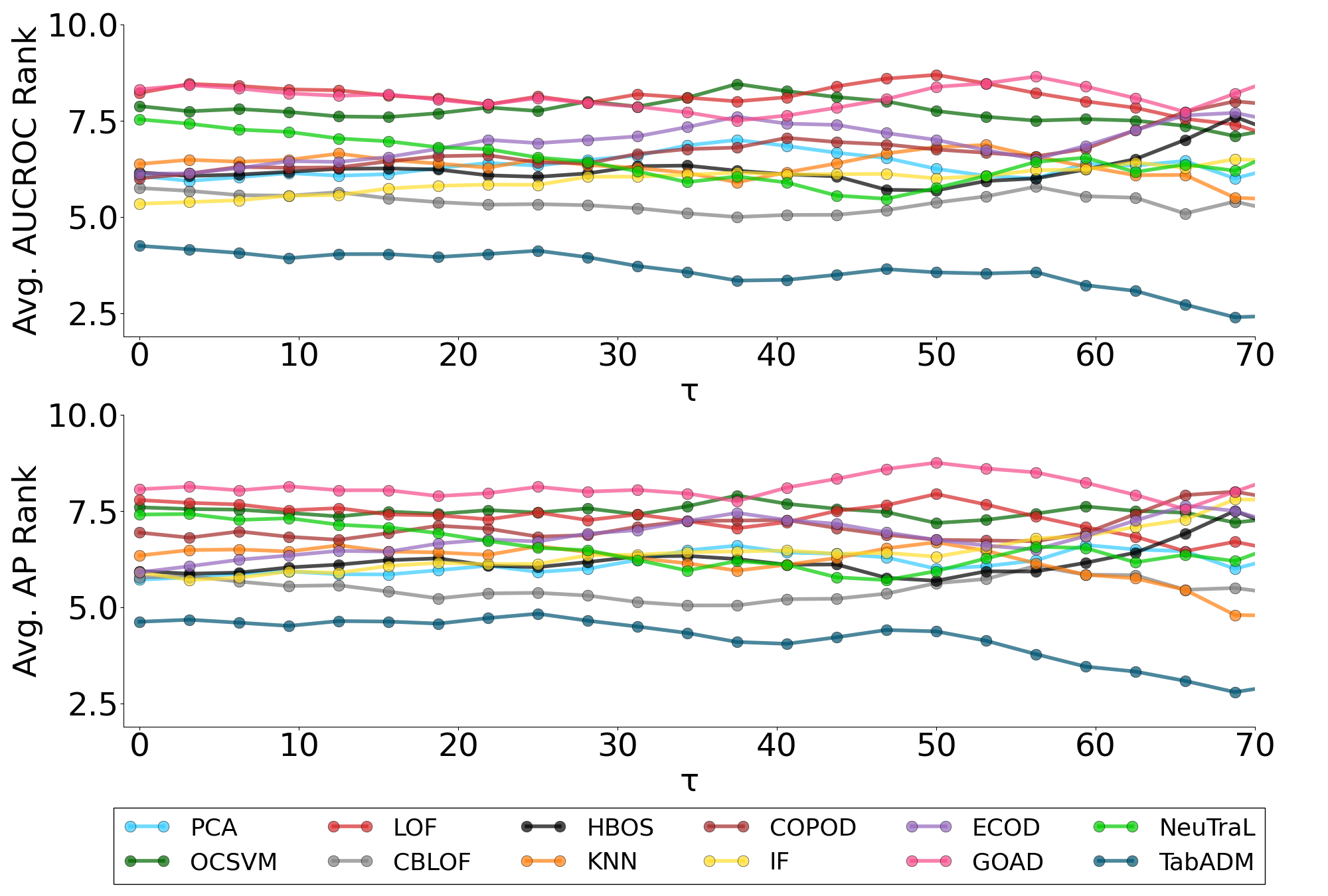}
\caption{Average AUCROC (top) and AP (bottom) rank per method as a function of $\tau$, where $\tau$ is the percentile value corresponding to the number of dimensions. For example, when $\tau=10$, we form a subgroup comprising the top 90\% of datasets with the highest number of variables and present the average ranks on this subgroup. We limit $\tau$ to a maximum of 70 to avoid an evaluation on a small subset of datasets.}
\label{figure:AUCROC and AP vs percentile}
\vskip -0.1 in
\end{figure}

The results for this part are presented in Tab. \ref{table:AUCROC all datasets} and \ref{table:AP all datasets}. The results of our proposed TabADM method demonstrate that, on average, it outperforms the other baselines in both AUCROC and AP scores, as well as in average rank, by a significant margin. Additionally, we observe that among the top 10 datasets with the highest dimensionality, TabADM achieves the highest AUCROC (AP) score in 5 (4) datasets. In light of this, we conduct a more in-depth analysis to evaluate the performance of the methods with respect to the dimensionality of the dataset. As illustrated in Fig. \ref{figure:AUCROC and AP vs percentile}, TabADM demonstrates consistently low average ranks across all percentile values in both AUCROC and AP scores. Additionally, it can be observed that as the percentile value increases, the performance of TabADM improves, and the gap to the rest grows. This suggests that our model is particularly well-suited for large datasets.

% All Datasets Table - AUCROC and Rank
\begin{table}[htb!]
\vskip -0.5 in
    \captionsetup{skip=1ex}
    \centering
    \caption{AUCROC scores for ADBench datasets.}
    \fontsize{7}{10}\selectfont
    \setlength{\tabcolsep}{4pt}
    \renewcommand{\arraystretch}{0.85}
    \begin{tabular}{l|ccccccccccc|c}
    \toprule
        \textbf{Dataset} & \textbf{PCA} & \textbf{OCSVM} & \textbf{LOF} & \textbf{CBLOF} & \textbf{HBOS} & \textbf{KNN} & \textbf{COPOD} & \textbf{IF} & \textbf{ECOD} & \textbf{GOAD} & \textbf{NeuTraL} & \textbf{TabADM}  \\
        \midrule
        Smtp & 81.01 & 74.94 & \textbf{93.07} & 83.82 & 81.48 & 90.61 & 90.95 & 89.19 & 87.30 & 87.61 & 79.15 & 86.21 \\ 
        Mammogra. & 87.05 & 83.72 & 73.15 & 80.15 & 83.91 & 82.69 & \textbf{89.00} & 84.56 & 88.96 & 57.27 & 52.89 & 82.79 \\
        Thyroid & 96.04 & 87.58 & 86.37 & 94.10 & 95.91 & 95.49 & 94.07 & 97.73 & \textbf{97.84} & 73.19 & 92.20 & 93.84 \\ 
        Glass & 74.52 & 55.27 & 66.02 & 85.59 & 83.23 & 88.39 & 78.06 & 80.65 & 76.02 & 64.41 & 49.25 & \textbf{88.71} \\ 
        Shuttle & 98.95 & 98.15 & 53.59 & 75.59 & 98.68 & 63.40 & 99.41 & \textbf{99.72} & 99.27 & 98.56 & 95.11 & 98.97 \\ 
        Donors & 82.85 & 73.11 & 61.65 & 65.36 & 71.88 & 62.68 & 81.41 & 77.76 & \textbf{88.80} & 30.90 & 50.26 & 72.48 \\ 
        PageBlocks & 90.59 & 89.30 & 72.67 & 87.37 & 81.14 & 81.42 & 87.74 & 89.26 & \textbf{91.55} & 80.24 & 83.98 & 90.83 \\ 
        Vowels & 62.54 & 58.31 & 94.76 & 90.69 & 68.74 & \textbf{98.23} & 52.45 & 79.17 & 61.43 & 93.46 & 45.94 & 96.80 \\ 
        Pendigits & 92.61 & 92.62 & 51.44 & 90.26 & 92.20 & 72.97 & 88.99 & \textbf{94.77} & 91.60 & 63.17 & 86.04 & 86.20 \\ 
        Hepatitis & 79.05 & 67.86 & 79.52 & 73.81 & 79.76 & 71.90 & \textbf{82.38} & 74.29 & 75.48 & 69.05 & 55.75 & 71.43 \\ 
        Cardio & \textbf{94.27} & 91.96 & 64.08 & 80.27 & 83.34 & 72.17 & 91.16 & 90.91 & 92.90 & 59.40 & 26.30 & 81.70 \\ 
        Cardiotocogr. & 75.82 & \textbf{79.19} & 58.43 & 65.97 & 60.51 & 54.63 & 66.80 & 68.85 & 79.13 & 39.79 & 77.29 & 60.97 \\ 
        Waveform & 62.08 & 51.86 & 70.51 & 71.39 & 67.84 & \textbf{72.21} & 71.94 & 69.17 & 59.38 & 70.80 & 66.16 & 71.40 \\ 
        Letter & 53.31 & 51.33 & 87.43 & 76.28 & 61.43 & 87.77 & 55.03 & 63.19 & 56.82 & 80.55 & 23.06 & \textbf{91.04} \\ 
        Ionosphere & 79.03 & 74.42 & 85.93 & 89.52 & 65.65 & 92.43 & 79.54 & 84.43 & 73.66 & 87.49	& 83.41 & \textbf{92.67} \\ 
        Landsat & 36.63 & 36.62 & 55.94 & 62.05 & 57.01 & 59.00 & 42.50 & 49.49 & 37.09 & 59.11 &	\textbf{64.16} & 58.61 \\ 
        Satellite & 60.34 & 59.97 & 55.17 & 72.97 & 75.27 & 65.53 & 63.47 & 71.38 & 58.51 & 64.90 & \textbf{77.25} & 72.53 \\ 
        Satimage-2 & 97.41 & 97.13 & 49.86 & 99.87 & 97.66 & 92.92 & 97.18 & 99.21 & 96.14 & 97.47 & \textbf{99.94} & 99.31 \\ 
        Celeba & \textbf{78.58} & 69.70 & 42.49 & 60.19 & 75.63 & 57.46 & 75.17 & 69.91 & 75.81 & 26.69 & 70.79 & 69.05 \\ 
        SpamBase & 56.36 & 54.96 & 45.14 & 57.78 & 65.69 & 54.82 & \textbf{69.11} & 61.37 & 66.00 & 38.26 & 40.02 & 58.93 \\ 
        Campaign & 73.80 & 66.77 & 56.57 & 65.85 & \textbf{79.61} & 72.40 & 78.74 & 71.12 & 77.51 & 41.93 & 74.77 & 72.36 \\ 
        Optdigits & 49.97 & 52.84 & 53.05 & 73.28 & \textbf{80.34} & 38.92 & 66.71 & 69.23 & 58.93 & 69.49 & 58.32 & 59.23 \\ 
        MNIST & 85.35 & 82.69 & 68.30 & 79.67 & 61.94 & 81.42 & 77.92 & 80.47 & 75.10 & 83.58 & \textbf{88.49} & 86.13 \\ 
        Musk & \textbf{100.00} & 81.19 & 37.80 & \textbf{100.00} & \textbf{100.00} & 70.93 & 94.71 & 99.97 & 95.58 & \textbf{100.00} & 76.15 & \textbf{100.00} \\ 
        Backdoor & 88.68 & 84.77 & 71.57 & 83.10 & 75.66 & 68.20 & 78.91 & 73.22 & 84.56 & 90.48 & 89.99 & \textbf{91.83} \\ 
        Speech & 51.39 & 50.99 & 53.99 & 51.17 & 51.92 & 52.26 & 53.13 & 53.50 & 51.46 & 46.74 & 40.19 & \textbf{54.54} \\ 
        Census & 65.58 & 52.77 & 48.05 & 58.85 & 62.75 & 64.12 & \textbf{66.62} & 60.77 & 65.41 & 53.62 & 50.03 & 64.99 \\ 
        FashionM. &	86.87 &	85.67 &	63.75 &	88.32 &	82.94 & 84.87 & 84.09 & 85.44 & 85.23 & 72.89 & 86.59 & \textbf{89.43} \\
        SVHN &	55.57 &	55.86 &	\textbf{65.79} &	59.70 &	51.03 &	62.51 &	51.50 &	55.97 &	53.09 &	58.93  & 64.10 & 62.75 \\
        Amazon &	52.61 &	52.87 &	55.39 &	53.33 &	53.14 &	\textbf{56.32} &	52.31 &	52.92 &	52.32 &	50.36 & 54.57 & 54.71 \\
        Yelp &	56.09 &	56.30 &	61.94 &	60.12 &	56.44 &	\textbf{62.71} &	55.33 & 56.93 &	55.60 &	50.36 & 59.58 & 58.77 \\
        InternetAds & 60.24 & 68.75 & 64.77 & 68.41 & 68.28 & 70.10 & 67.51 & 67.90 & 67.57 & 41.82 &	46.10 & \textbf{76.48} \\ \hline
        AVG & 73.91 & 69.98 & 64.01	& 75.15	& 74.09 & 71.92 &	74.49 &	75.70 &	74.25 & 65.70 &	65.87 & \textbf{77.99} \\ 
        AVG Rank & 6.06 &	7.88 &	8.22 &	5.75 &	6.16 &	6.38 &	6.00 &	5.34 &	6.13 &	8.31 &	7.53 &	\textbf{4.25} \\ \hline
    \end{tabular}
     %The datasets are arranged in ascending order of the number of dimensions.}
    \label{table:AUCROC all datasets}
    %\end{center}
    \vskip -0.4 in
\end{table}

% All Datasets Table - AP and Rank
\begin{table}[H]
    \captionsetup{skip=1ex}
    \centering
    \caption{AP scores for ADBench datasets.}% The datasets are arranged in ascending order of the number of dimensions.}
    \label{table:AP all datasets}
    \fontsize{7}{10}\selectfont
    \setlength{\tabcolsep}{4pt}
    \renewcommand{\arraystretch}{0.85}
    \begin{tabular}{l|ccccccccccc|c}
    \toprule
        \textbf{Dataset} & \textbf{PCA} & \textbf{OCSVM} & \textbf{LOF} & \textbf{CBLOF} & \textbf{HBOS} & \textbf{KNN} & \textbf{COPOD} & \textbf{IF} & \textbf{ECOD} & \textbf{GOAD} & \textbf{NeuTraL} & \textbf{TabADM} \\
        \midrule
        Smtp & 44.75 & 5.35 & 1.43 & 44.42 & 8.70 & 48.38 & 0.46 & 0.44 & \textbf{62.09} & 41.92 & 31.36 & 46.10 \\ 
        Mammogra. & 19.93 & 11.84 & 10.33 & 7.96 & 19.35 & 16.10 & 40.23 & 19.50 & \textbf{40.67} & 5.84 &	3.04 & 15.68 \\ 
        Thyroid & 42.26 & 19.49 & 16.87 & 26.16 & 52.55 & 29.88 & 19.41 & \textbf{60.16} & 49.60 & 35.65 & 34.44 & 31.19 \\ 
        Glass & 20.65 & 16.81 & 22.53 & 25.28 & 25.42 & 26.82 & 18.97 & 21.59 & 25.68 & 16.67 & 7.80 & \textbf{39.06} \\ 
        Shuttle & 90.35 & 94.69 & 9.68 & 40.51 & 95.07 & 16.42 & 95.56 & \textbf{97.92} & 89.81 & 85.60 & 52.89 & 92.52 \\ 
        Donors & 16.68 & 10.42 & 10.50 & 8.25 & 12.24 & 10.64 & 20.76 & 13.07 & \textbf{26.33} & 5.13 &	5.58 & 11.59 \\ 
        PageBlocks & 52.82 & 51.88 & 35.77 & 56.64 & 35.34 & 45.44 & 37.42 & 45.78 & 52.34 & 46.15 & 29.95 & \textbf{57.27} \\ 
        Vowels & 8.54 & 9.19 & 43.12 & 23.99 & 11.64 & \textbf{67.26} & 4.05 & 15.79 & 10.46 & 39.49 &	5.28 & 64.12 \\ 
        Pendigits & 19.86 & 19.51 & 3.88 & 16.37 & 24.16 & 6.23 & 16.54 & \textbf{28.71} & 25.19 & 5.92 &	14.95 & 9.83 \\ 
        Hepatitis & \textbf{50.94} & 28.67 & 37.08 & 31.76 & 43.31 & 29.73 & 50.84 & 34.82 & 36.97 & 33.23	& 22.54 & 32.30 \\ 
        Cardio & \textbf{57.18} & 53.33 & 18.25 & 36.56 & 42.68 & 31.25 & 52.56 & 48.48 & 53.55 & 29.43 & 7.95 & 31.77 \\ 
        Cardiotocogr. & 48.48 & 54.43 & 30.42 & 44.03 & 39.37 & 33.88 & 40.78 & 43.03 & 51.65 & 24.40 & \textbf{61.81} & 37.07 \\ 
        Waveform & 4.80 & 3.94 & 9.22 & 15.11 & 5.08 & 10.77 & 5.67 & 6.21 & 4.47 & \textbf{44.14} & 5.32 & 6.26 \\ 
        Letter & 10.16 & 8.62 & 43.76 & 19.55 & 10.32 & 34.73 & 7.36 & 9.91 & 8.63 & 30.47 & 4.25 & \textbf{49.66} \\ 
        Ionosphere & 73.13 & 72.76 & 81.22 & 87.74 & 46.00 & 92.36 & 68.86 & 79.49 & 65.74 & 85.59 & 75.54 & \textbf{92.37} \\ 
        Landsat & 16.43 & 16.33 & 26.49 & \textbf{29.36} & 23.26 & 25.95 & 17.89 & 20.12 & 16.61 & 24.89 &	26.72 & 24.95 \\ 
        Satellite & 60.80 & 58.98 & 38.45 & 62.19 & 67.70 & 50.55 & 57.18 & 66.20 & 52.65 & 43.94 & \textbf{72.87} & 56.15 \\ 
        Satimage-2 & 87.19 & 86.03 & 3.37 & \textbf{97.32} & 80.89 & 34.83 & 79.94 & 91.31 & 68.14 & 43.72 & 96.35 & 62.75 \\ 
        Celeba & \textbf{11.28} & 7.41 & 1.78 & 3.30 & 9.77 & 2.69 & 9.63 & 7.08 & 9.86 & 1.39	& 4.62 & 4.37 \\ 
        SpamBase & 42.21 & 41.59 & 35.91 & 43.27 & 50.36 & 42.70 & \textbf{54.74} & 50.14 & 52.21 & 34.74 & 35.71 & 43.69 \\ 
        Campaign & 28.97 & 28.58 & 12.12 & 21.12 & \textbf{37.45} & 26.43 & 37.01 & 30.23 & 35.57 & 10.46 & 31.10 & 29.52 \\ 
        Optdigits & 2.67 & 2.90 & 5.00 & 4.99 & \textbf{11.01} & 2.54 & 4.21 & 5.19 & 3.40 & 5.01 & 3.70 & 4.22 \\ 
        MNIST & 39.63 & 32.97 & 22.73 & 30.91 & 13.41 & 38.31 & 22.37 & 28.43 & 18.27 & 38.35 & 41.96 & \textbf{44.17} \\ 
        Musk & \textbf{100.00} & 10.47 & 3.57 & \textbf{100.00} & \textbf{100.00} & 10.78 & 35.61 & 99.80 & 48.73 & \textbf{100.00} & 6.98 & \textbf{100.00} \\ 
        Backdoor & 52.13 & 8.62 & 20.69 & 7.19 & 5.47 & 30.55 & 6.92 & 3.91 & 9.25 & \textbf{55.48} & 36.38 & 41.29 \\ 
        Speech & 2.98 & 3.07 & 3.90 & 3.08 & 3.19 & 3.37 & 3.03 & 2.84 & 3.28 & 2.38 & 1.48 & \textbf{3.99} \\ 
        Census & 8.76 & 6.27 & 5.51 & 6.94 & 7.88 & 8.16 & \textbf{8.98} & 7.33 & 8.58 & 6.54 & 5.78 & 8.51 \\ 
        FashionM. &	26.42 &	24.90 &	11.27 &	28.40 &	23.65 & 23.96 & 24.21 & 23.04 & 24.84 & 17.74 & 31.89 & \textbf{37.82} \\
        SVHN &	7.03 &	7.02 &	\textbf{9.83} &	8.12 &	6.00 &	8.81 &	6.16 &	7.04 &	6.31 & 7.84	& 8.94 & 8.49 \\
        Amazon &	5.43 &	5.44 &	5.69 &	5.48 &	5.48 & \textbf{5.79} &	5.40 &	5.49 &	5.35 & 5.26 & 5.63 &	5.60 \\
        Yelp &	6.01 &	6.11 &	7.56 &	6.50 &	6.15 &	\textbf{7.57} &	6.04 &	6.17 &	5.84 &	5.01 & 6.85 & 6.55 \\
        InternetAds & 27.54 & \textbf{53.29} & 37.81 & 53.15 & 52.33 & 41.63 & 50.35 & 50.04 & 50.46 & 53.05 & 17.89 & 53.05 \\ \hline
        AVG  & 33.94 &	26.90 &	19.55 &	31.11 &	30.48 &	27.02 &	28.41 &	32.16 & 	31.95 &	29.70 &	24.92 & \textbf{36.00} \\ 
        AVG Rank  & 5.72 &	7.59 &	7.78 &	5.78 &	5.94 &	6.34 &	6.94 &	5.91 &	5.91 &	8.06 &	7.41 &	\textbf{4.63} \\ \hline
    \end{tabular}

   %\end{center}
\end{table}

% AUC and AP vs Contamination 3 graphs

%% Second part

In the \textbf{second part}, we randomly divide \textit{Satellite}, \textit{Cardiotocography}, and \textit{SpamBase} datasets into training and test sets using a 70-30 split. Then, we create 11 sub-training sets with varying contamination ratios from 0\% to 10\%. In addition, we randomly fix a 10\% contamination ratio in the test set. We repeat this process 5
times and plot the average AUCROC and AP scores as a function of contamination ratios for each dataset. 

\begin{figure}[H]
  \vskip -0.2 in
\centering
\includegraphics[width=1\textwidth]{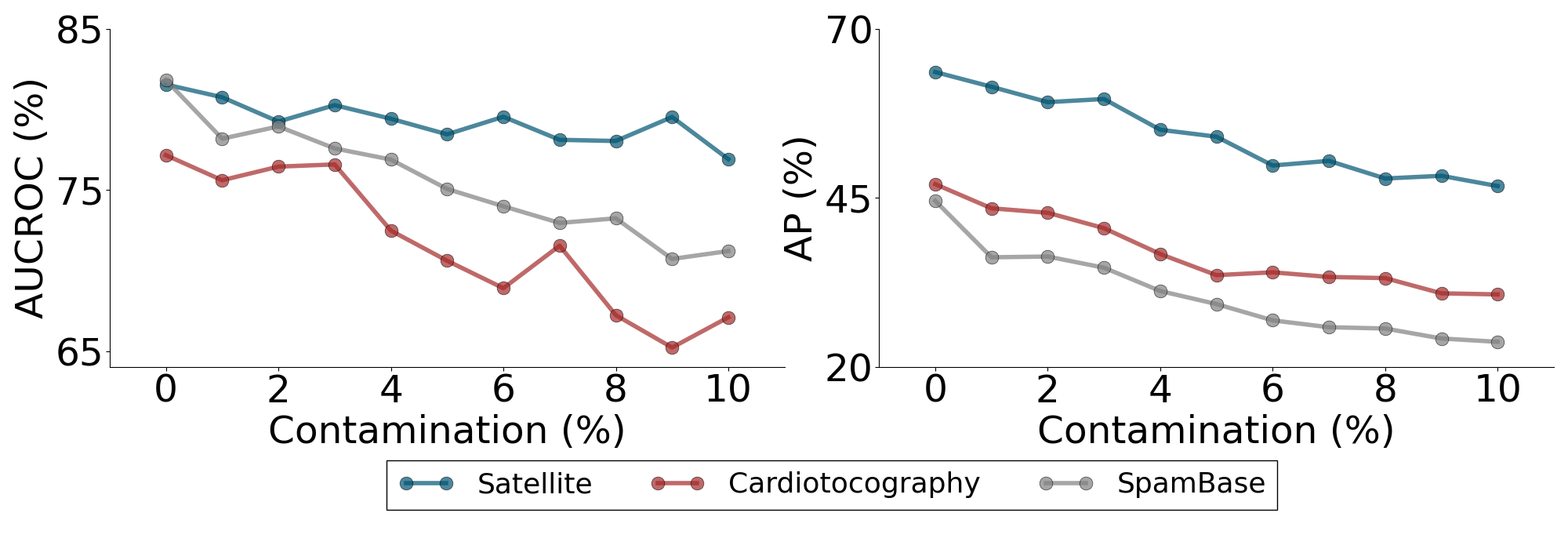}
\caption{AUCROC (left) and AP (right) scores for the \textit{Satellite}, \textit{Cardiotocography}, and \textit{SpamBase} datasets decrease as the contamination percentage increases. This is due to the increasing influence of anomalous samples on the overall probability distribution learned by the model.}
\label{figure:AUC and AP vs contamination}
\end{figure}

The results in this part are shown in Fig. \ref{figure:AUC and AP vs contamination}. As the level of contamination in the training set increases, there is a decline in both the AUCROC and AP scores. This can be attributed to the fact that the model learns the anomalous samples in addition to the inlier samples. As a result, the ability of our model to accurately distinguish between inliers and outliers is hindered, leading to a decrease in performance.

%% Third part
In the \textbf{third part}, we investigate the impact of different training hyperparameters on the performance of our model. We examine the relationship between the AUCROC and AP scores and the number of training iterations for \textit{Landsat}, \textit{Letter}, and \textit{Musk}. In addition, we investigate the influence of the number of rejections samples (m) on the performance. As in previous parts, we use a 70-30 train-test random split over five times and report the average AUCROC and AP scores for $m=0,1,4,7$. 

\begin{figure}[H]
\centering
  \vskip -0.0 in
\includegraphics[width=1\textwidth]{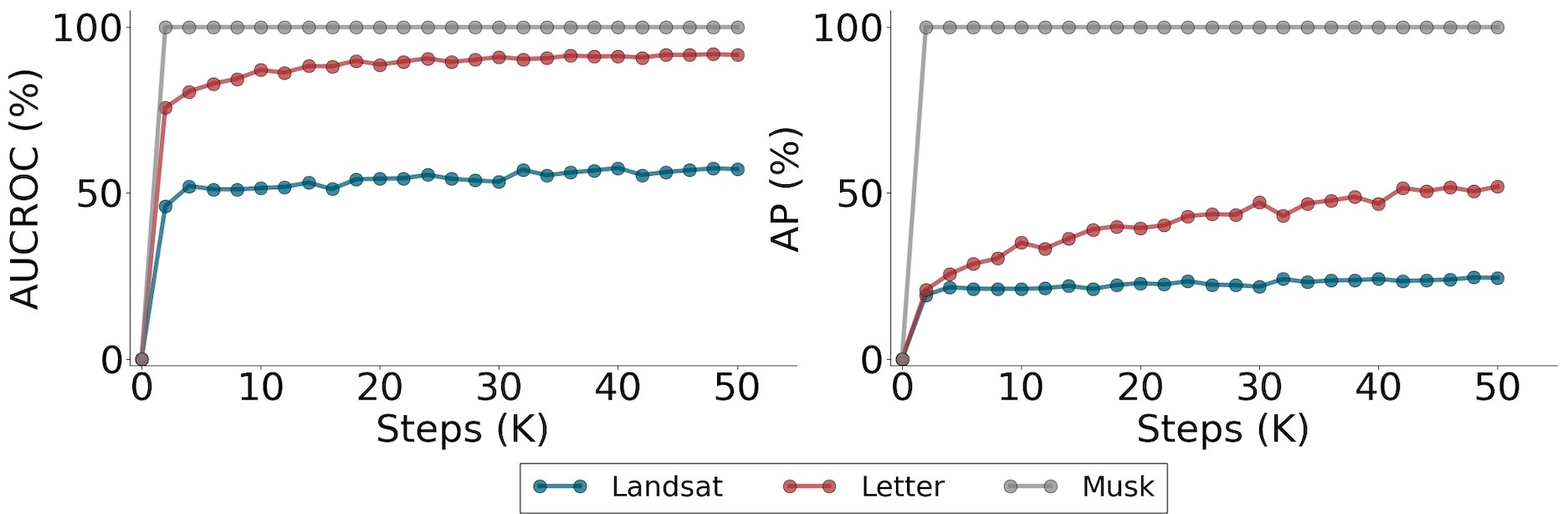}
\caption{AUCROC (left) and AP (right) scores for the \textit{Landsat}, \textit{Letter}, and \textit{Mask} datasets as functions of training steps.}
\label{figure:AUC and AP vs steps}
\end{figure}

Fig. \ref{figure:AUC and AP vs steps} and Tab. \ref{table:AUC and AP with or without removal} present the results for this part. As shown in Fig. \ref{figure:AUC and AP vs steps}, as the number of training steps increases, the performance of all datasets improves. However, the improvement rate varies among different datasets. Tab. \ref{table:AUC and AP with or without removal} demonstrates that excluding the sample with the highest loss in a batch during training ($m=1$) leads to the highest average scores. This indicates that the model is more robust to anomalies, resulting in better modeling of the normal underlying distribution and, consequently, improved overall performance.

\begin{table}[htbp]
\centering
  \vskip -0.2 in
\captionsetup{skip=1ex}
\caption{Comparison of AUCROC (left) and AP (right) scores for the \textit{Landset}, \textit{Letter} and \textit{Musk} datasets for different values of rejection samples $m$ from batch of size 8.}
\label{table:AUC and AP with or without removal}
\begin{tabular}{lccccccccc}
\toprule
& \multicolumn{4}{c}{\textbf{AUCROC} (\%)}& \multicolumn{4}{c}{\textbf{AP} (\%)} \\ 
\cmidrule(lr){2-5} \cmidrule(lr){6-9}
Dataset & m=0 & m=1 & m=4 & m=7 & m=0 & m=1 & m=4   & m=7 \\ 
\midrule
Landsat	& 56.78 & \textbf{58.61} & 55.78 & 54.88 & 24.41 & \textbf{24.95} & 22.73 & 22.67  \\ 
Letter     & \textbf{91.28} & 91.04 & 90.75 & 81.06 &	48.24 & \textbf{49.66} & 39.81 & 22.00 \\ 
Musk     & 	99.10	& \textbf{100.00} & \textbf{100.00} & \textbf{100.00} &	72.56 & \textbf{100.00} & \textbf{100.00} & \textbf{100.00} \\ 
\bottomrule
\end{tabular}
\end{table}

\section{Conclusion and Future Work} \label{CONCLUSION}
In this paper, we introduce a novel unsupervised outlier detection method, TabADM, which utilizes the diffusion models technique to estimate the probability distribution of the data. It then assigns outlier scores to unseen samples based on their probability of being generated from the model. In addition, a rejection scheme is introduced to enhance performance when outliers are present in the data. TabADM exhibits strong training stability and alleviates the need for hyperparameter tuning. Furthermore, it demonstrates exceptional performance in high-dimensional datasets, surpassing other SOTA methods.  

TabADM has certain drawbacks, including long training and inference times compared to other methods and a lack of interpretability.
Future work could focus on improving these drawbacks. For example, the inference time can be reduced by decreasing the number of diffusion steps used per sample, although this may impact performance. Additionally, efforts could be made to enhance interpretability. This could be achieved through simple measures such as identifying which features contribute most significantly to the total loss, as well as more complex measures such as identifying common feature patterns in the data that may serve as indicators for abnormality. Another possible future research direction would be to extend the capabilities of TabADM such as enabling it to handle missing feature values.
%#, as well as categorical or ordinal features in the data.

\section*{Acknowledgments}
Funding: This research was partially supported by the Israel Science Foundation (ISF, 1556/17, 1873/21), Israel Ministry of Science Technology and Space 3-16414, 3-14481,3-17927) and Magneton Playtika.4758/2

\bibliographystyle{plain}
\bibliography{paper}

\appendix
\newpage
\section*{Appendix}
\section{Defualt hyperparameters}
Tab. \ref{table:hyperparameters list} presents the default hyperparameters employed for the training procedure in our experiments. While most hyperparameters remain constant, the learning rate and the size of the fully connected (FC) layers are determined by the number of dimensions ($d$) of the training dataset. 
% Hyperparameters table
\begin{table}[H]
    \captionsetup{skip=1ex}
    \centering
    \caption{Default hyperparameters for TabADM training procedure.}
    \label{table:hyperparameters list}
    \begin{tabular}{l|ccc}
    \toprule
        \textbf{\# Dims. in dataset (d)} & \boldmath{$d\leq100$} & \boldmath{ $100< d\leq1000$} & \boldmath{$1000 < d \leq 2000$} \\ \midrule
        FC layers size & 512 & 1024 & 2048 \\ 
        Learning rate & $1X10^{-3}$ & $2X10^{-4}$& $2X10^{-4}$ \\ 
        Weight decay & $1X10^{-4}$ & $1X10^{-4}$ & $1X10^{-4}$ \\ 
        Time embedding size & 64 & 64 & 64 \\ 
        Batch size (k) & 8 & 8 & 8 \\ 
        Rejection samples (m) & 1 & 1 & 1 \\ 
        Train steps (e) & 50K & 50K & 50K \\ 
        Noise schedule & linear & linear & linear  \\ 
        Diffusion timesteps (T) & 100 & 100 & 100 \\ 
        Loss type & Simplified MSE of $\epsilon$ & Simplified MSE of $\epsilon$ & Simplified MSE of $\epsilon$ \\ \hline
    \end{tabular}
\end{table}
\newpage
\section{Evaluation datasets} Tab. \ref{table:datasets list} lists the ADBench \cite{ADBench} datasets used for evaluation. To ensure a fair comparison between the baselines, we selected diverse datasets in terms of their dimensions ($d$), number of samples ($n$), and anomaly rates. 

The list contains 32 datasets. Among them, 28 real-world tabular datasets and the other 4 are extracted feature embedding representations of pre-trained models from CV and NLP fields. The CV datasets include \textit{FashionMNIST} and \textit{SVHN}, and we randomly selected the first class (out of 10 classes) for testing. In addition, each dataset has two versions: the features of the first version are based on a pre-trained \textit{ResNet} model with 512 dimensions, while the features of the second version are based on a pre-trained \textit{ViT} model with 1000 dimensions. Similarly, the NLP datasets include \textit{Amazon} and \textit{Yelp}, each with two versions. One is based on a pre-trained \textit{BERT} model, and the other is based on a pre-trained \textit{RoBERTa} model. Both versions have 768 dimensions. For the CV and NLP datasets, we reported the average AUCROC and AP scores across the two versions. Lastly, due to convergence failure in some of the baselines, we randomly sub-sampled \textit{Census} to 50,000 samples in a stratified way to preserve the original anomaly rate post-truncation. 

% Datasets List
\begin{table}[H]
    \captionsetup{skip=1ex}
    \centering
    \caption{List of ADBench datasets used for evaluation.}
    \label{table:datasets list}
    \begin{tabular}{c|ccc}
    \toprule
        \textbf{Dataset} & \textbf{\# Dims. ($d$)} & \textbf{\# Samp. ($n$)} & \textbf{Anomaly rate} (\%) \\
        \midrule
        Smtp & 3 & 95156 & 0.03 \\
        Mammography & 6 & 11183 & 2.32 \\
        Thyroid & 6 & 3772 & 2.47 \\
        Glass & 7 & 214 & 4.21 \\
        Shuttle & 9 & 49097 & 7.15 \\
        Donors & 10 & 619326 & 5.93 \\
        PageBlocks & 10 & 5393 & 9.46 \\
        Vowels & 12 & 1456 & 3.43 \\
        Pendigits & 16 & 6870 & 2.27 \\
        Hepatitis & 19 & 80 & 16.25 \\
        Cardio & 21 & 1831 & 9.61 \\
        Cardiotocography & 21 & 2114 & 22.04 \\
        Waveform & 21 & 3443 & 2.90 \\
        Letter & 32 & 1600 & 6.25 \\
        Ionosphere & 33 & 351 & 35.90 \\
        Landsat & 36 & 6435 & 20.71 \\
        Satellite & 36 & 6435 & 31.64 \\
        Satimage-2 & 36 & 5803 & 1.22 \\
        Celeba & 39 & 202599 & 2.24 \\
        SpamBase & 57 & 4207 & 39.91 \\
        Campaign & 62 & 41188 & 11.27 \\
        Optdigits & 64 & 5216 & 2.88 \\
        MNIST & 100 & 7603 & 9.21 \\
        Musk & 166 & 3062 & 3.17 \\
        Backdoor & 196 & 95329 & 2.44 \\
        Speech & 400 & 3686 & 1.65 \\
        Census & 500 & 50000 & 6.20 \\
        FashionMNIST & 512/1000 & 6315 & 5.00 \\
        SVHN & 512/1000 & 5208 & 5.00 \\
        Amazon & 768/768 & 10000 & 5.00 \\
        Yelp & 768/768 & 5000 & 5.00 \\
        InternetAds & 1555 & 1966 & 18.72 \\ \hline
    \end{tabular}
%\vskip -0.2 in
\end{table}

\end{document}